\documentclass[sigconf]{acmart}

\usepackage{bm} 
\usepackage{multirow}
\usepackage{array}
\usepackage{graphicx}
\usepackage{amsmath}
\usepackage{amsthm}
\usepackage{booktabs}
\usepackage{algorithm}
\usepackage{algpseudocode}
\usepackage{amsfonts}
\usepackage{amsmath}
\usepackage{multirow}
\usepackage{tabularx}
\usepackage{booktabs}
\usepackage{subfigure}
\usepackage{diagbox}
\usepackage{pifont}

\newcommand{\cmark}{\ding{51}}%
\newcommand{\xmark}{\ding{55}}%

\newcommand{\code}[0]{\url{https://github.com/EdisonLeeeee/STEP}}

\AtBeginDocument{%
  \providecommand\BibTeX{{%
    \normalfont B\kern-0.5em{\scshape i\kern-0.25em b}\kern-0.8em\TeX}}}

\setcopyright{acmcopyright}
\copyrightyear{2018}
\acmYear{2018}
\acmDOI{XXXXXXX.XXXXXXX}

\acmConference[Conference acronym 'XX]{Make sure to enter the correct
  conference title from your rights confirmation emai}{June 03--05,
  2018}{Woodstock, NY}
\acmPrice{15.00}
\acmISBN{978-1-4503-XXXX-X/18/06}

\begin{document}

\title{Less Can Be More: Unsupervised Graph Pruning for Large-scale Dynamic Graphs}


\author{Jintang Li}
\authornote{Both authors contributed equally to this research.}
\affiliation{\institution{Sun Yat-sen University}
	\country{}}
\email{lijt55@mail2.sysu.edu.cn}

\author{Sheng Tian}
\authornotemark[1]
\affiliation{\institution{Ant Group}
	\country{}}
\email{tiansheng.ts@antgroup.com}

\author{Ruofan Wu}
\affiliation{\institution{Ant Group}
	\country{}}
\email{ruofan.wrf@antgroup.com}

\author{Liang Zhu}
\affiliation{\institution{Ant Group}
	\country{}}
\email{tailiang.zl@antgroup.com}

\author{Wenlong Zhao}
\affiliation{\institution{Ant Group}
	\country{}}
\email{chicheng.zwl@antgroup.com}

\author{Changhua Meng}
\affiliation{\institution{Ant Group}
	\country{}}
\email{changhua.mch@antgroup.com}

\author{Liang Chen}
\affiliation{\institution{Sun Yat-sen University}
	\country{}}
\email{chenliang6@mail.sysu.edu.cn}

\author{Zibin Zheng}
\affiliation{\institution{Sun Yat-sen University}
	\country{}}
\email{zhzibin@mail.sysu.edu.cn}

\author{Hongzhi Yin}
\affiliation{\institution{The University of Queensland}
	\country{}}
\email{db.hongzhi@gmail.com}

\renewcommand{\shortauthors}{Li and Tian, et al.}

\begin{abstract}
	The prevalence of large-scale graphs poses great challenges in time and storage for training and deploying graph neural networks (GNNs). Several recent works have explored solutions for pruning the large original graph into a small and highly-informative one, such that training and inference on the pruned and large graphs have comparable performance. Although empirically effective, current researches focus on static or non-temporal graphs, which are not directly applicable to dynamic scenarios. In addition, they require labels as ground truth to learn the informative structure, limiting their applicability to new problem domains where labels are hard to obtain. To solve the dilemma, we propose and study the problem of \textit{unsupervised graph pruning} on dynamic graphs. We approach the problem by our proposed STEP, a self-supervised temporal pruning framework that learns to remove potentially redundant edges from input dynamic graphs. From a technical and industrial viewpoint, our method overcomes the trade-offs between the performance and the time \& memory overheads. Our results on three real-world datasets demonstrate the advantages on improving the efficacy, robustness, and efficiency of GNNs on dynamic node classification tasks. Most notably, STEP is able to prune more than 50\% of edges on a million-scale industrial graph Alipay (7M nodes, 21M edges) while approximating up to 98\% of the original performance. Code is available at \code.
\end{abstract}



\maketitle

\section{Introduction}
\label{sec:intro}

\begin{figure}[t]
	\centering
	\includegraphics[width=\linewidth]{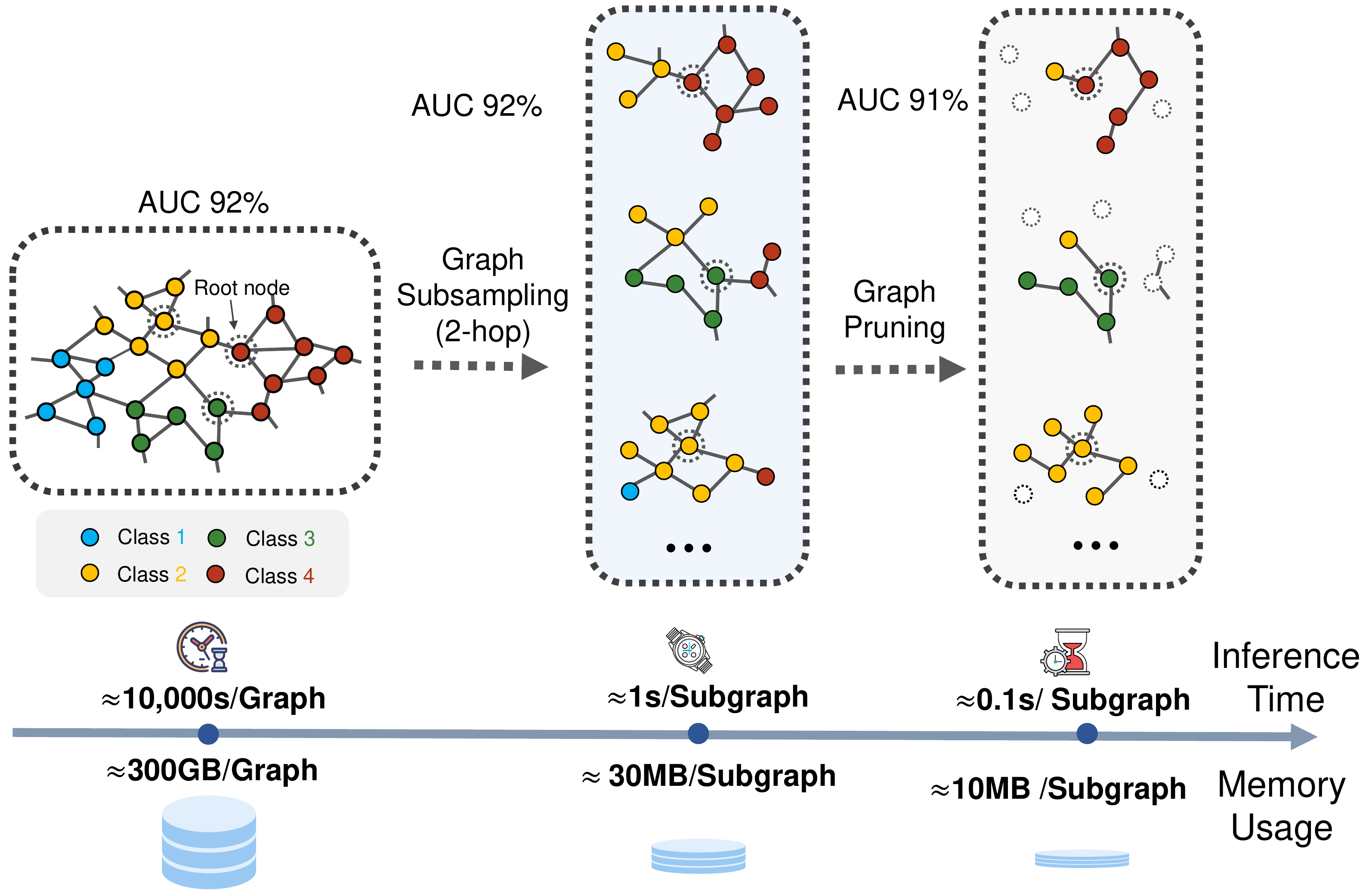}
	\caption{An illustrative example on graph subsampling and graph pruning. For each training node (dashed circle), graph subsampling samples a fixed-size set of neighbors as the computation graph, while graph pruning further removes redundant edges to improve inference and storage efficiency without significantly sacrificing the downstream performance.}
	\label{fig:example}
\end{figure}

Graphs are fundamental data structures that represent pairwise relationships between entities in various scientific and commercial fields~\cite{gnn_survey,maskgae}.
As a generalization of deep neural networks, graph neural networks (GNNs) have emerged as demonstrably powerful models for learning on graph-structured data.
The past few years have witnessed the success of GNNs in a variety of graph-based applications, such as recommender systems~\cite{DBLP:conf/wsdm/GaoW0022,pinsage}, biomedical discovery~\cite{DBLP:conf/nips/DaiLCDS19}, and social network analysis~\cite{qiu2018deepinf}.

Despite the progress, building and deploying GNNs to handle real-world graph data poses several fundamental challenges. The first and most prominent challenge is to scale GNNs to deal with large-scale graphs since canonical message passing in GNNs performed over the whole graph would lead to expensive computation and memory costs.
However, industry problems of real-world interest often involve giant graphs with billions of nodes and tens of billions of edges.
For example, the financial graph in Alipay\footnote{\url{https://global.alipay.com/platform/site/ihome}} has enormous transfer transactions carried out between users, in which billions of cases of embezzlement, fraud, money laundering, and gambling occur dynamically every day.
Their massive size requires heavy floating point operations and large memory footprints, making them infeasible to be stored in the memory of a single machine and posing challenges to deploying GNNs on cheap devices.

In literature, considerable research efforts have been devoted to improving the scalability and efficiency of GNNs in dealing with large-scale graphs~\cite{DBLP:journals/pvldb/ZhangYWSLW021,agl,DBLP:journals/pvldb/ZhouSZKP21}.
As the main drawback that makes GNNs suffer from scalability issues is the need to operate on the entire graph, one practical approach is to sample smaller subgraphs and fed them to GNNs via mini-batched manner, also known as \textit{graph subsampling}. One prominent example is PinSAGE~\cite{pinsage}, a web-scale recommender system based on GNN architectures, which is successfully applied to Pinterest data with billions of nodes and tens of billions of edges.
However, there still exist some limitations for graph subsampling.
First, it mainly solves the scalability problem during training, but additionally leads to more memory overheads for offline storage of the sampled subgraphs. Second, subsampling can make some nodes appear multiple times, which would potentially introduce a lot of redundant information (or even bias) covering the sampled subgraphs.
Third, a ``clean'' structure of the sampled subgraph is not guaranteed in most cases. If task-irrelevant information (e.g., noise~\cite{learn_to_drop,nrgnn} or adversarial perturbations~\cite{median_gcn,chen2020survey}) is mixed into nodes' neighborhood, it may also ``dilute'' the truly useful signals from a small set of close neighbors.

One way to rectify the aforementioned issues is graph pruning~\cite{dropedge,neural_sparse,pgexplainer}, a new yet promising technology that has paved the way to meet the challenges of reliable graph learning at scale. By properly removing redundant or even noisy edges from the input graphs, graph pruning can not only boost the performance of graph algorithms (e.g., GNNs), but also facilitate storage, training, and inference efficiency for graph-based analysis tasks. In addition, graph pruning can also work with graph subsampling to further address its limitations. We provide sketched plots for visual illustration in Figure~\ref{fig:example}. Graph pruning is critical for large-scale graph applications. However, current approaches suffer from two fundamental challenges when dealing with industrial problems.

\textbf{(i) Tackling dynamic graphs.}
Real-life industrial applications often involve graphs that are inherently dynamic, with changes in the existence of nodes and edges over time.
While temporal changes (dynamics) play an essential role in capturing evolutionary patterns of topology structures, most of the research in the field is designed for static graphs and uses static computation strategies.
New methods are needed that can handle streaming graph events and prune new arriving edges in a dynamically evolving graph.

\textbf{(ii) Tackling unsupervised problem.}
Previous work on graph pruning typically requires access to abundant labels to remove edges which would potentially hinder the downstream performance. In many situations, however, label annotations can be prohibitively expensive or even impossible to collect, making them less applicable for these problems. The challenge here is to learn the pruning algorithm without relying on label annotations and feedback signals from a specific downstream task.

\begin{table}[t]
	\centering
	\caption{Comparison of STEP and other pruning methods.}\label{tab:comparison}
	\resizebox{\linewidth}{!}
	{\begin{tabular}{l|cccc}
			\toprule
			\textbf{Method}                   & \textbf{Dynamic} & \textbf{Unsupervised} & \textbf{Graph-less} & \textbf{Learnable} \\
			\midrule
			DropEdge~\cite{dropedge}          & \xmark           & \cmark                & \cmark              & \xmark             \\
			NeuralSparse~\cite{neural_sparse} & \xmark           & \xmark                & \xmark              & \cmark             \\
			PGExplainer~\cite{pgexplainer}    & \xmark           & \xmark                & \xmark              & \cmark             \\
			\midrule
			STEP (ours)                       & \cmark           & \cmark                & \cmark              & \cmark             \\
			\bottomrule
		\end{tabular}
	}
\end{table}

To our best knowledge, little effort has done to fully address above serious problems of graph pruning.  In light of the challenges, we propose a \underline{\textbf{S}}elf-supervised \underline{\textbf{TE}}mporal \underline{\textbf{P}}runing (STEP for abbreviation) framework, for pruning large-scale dynamic graphs of potentially redundant edges. As listed in Table~\ref{tab:comparison}, our proposed framework enjoy several advantages over other widely-adopted pruning methods: (i) STEP is capable of managing streaming events on dynamic graphs timely and inductively (\textbf{dynamic}). (ii) STEP additionally introduces a compact network that takes each individual edge as input for pruning without explicit message passing at \textit{inference} (\textbf{graph-less}). (iii) Instead of learning by feedback signals from specific downstream supervisions, STEP is able to learn node/edge representations and adaptively discover the underlying graph structure with the aid of self-supervised contrastive learning (\textbf{unsupervised and learnable}).

We highlight the key contributions of this work as follows:
\begin{itemize}
	\item We propose STEP, a simple yet effective pruning framework for dynamic graphs, which handles the problem of pruning dynamic graphs represented with streaming events timely and unsupervisedly.
	\item Our proposed STEP is an entirely unsupervised framework, which learns the underlying dynamic graph structure without need of task-specific feedback signals (e.g., labels).
	\item We present numerical results to demonstrate the superiority of the proposed framework over several pruning baselines. STEP is beneficial for GNN alternatives in terms of efficacy, robustness, and efficiency on large-scale dynamic graphs.
\end{itemize}

\section{Related Work}

\label{sec:related_work}

\subsection{Dynamic Graph Representation Learning}\label{sec:related_dgrl}

Over the past few years, learning representations for dynamic graphs has attracted considerable research effort, where most works are limited to the setting of discrete-time dynamic graphs (DTDG) represented as a sequence of snapshots~\cite{EvolveGCN,tNodeEmbed}. Such approaches are often insufficient for real-world systems as graphs in most cases are represented with a sequence of continuous events (i.e., nodes and edges can appear and disappear at any time). Until recently, several architectures for continuous-time dynamic graphs (CTDG) have been proposed~\cite{dynamic_graph_survey}. JODIE~\cite{jodie} uses a recurrent architecture to learn embedding trajectories of each node.
DyREP~\cite{dyrep} utilizes a parameterized temporal point process model to capture evolution patterns over historical structure.
TGAT~\cite{tgat} proposes to use self-attention mechanisms to handle the streaming graph events inductively and has reported state-of-the-art performances in graph-based learning tasks. We refer readers to \cite{dynamic_graph_survey} and the references therein for more thorough reviews of dynamic representation learning on graphs.
\subsection{Graph Self-supervised Learning}\label{sec:related_gssl}
Recently, graph self-supervised learning has become a new trend in deep graph learning and attracted considerable attention from both academics and the industry.
In self-supervised learning, models are learned by solving a series of well-designed auxiliary tasks, in which the supervision signals are acquired from the data itself automatically without the need for manual annotations~\cite{DBLP:journals/corr/abs-2103-00111,maskgae}.
Among contemporary approaches, contrast learning is one of the most successful self-supervised learning paradigms on graph data~\cite{velickovic2019deep,DBLP:conf/aaai/ZengX21,xu2021infogcl}. The primary goal of contrastive learning is to encourage two augmentation views of graphs to have more similar representations (typically via mutual information maximization~\cite{info_nce}).
Contrastive learning applied to self-supervised graph representation learning has led to state-of-the-art performance in a wide range of graph-based analysis tasks. Nevertheless, relatively little effort has been made toward graph pruning in dynamic settings.

\begin{figure*}[t]
	\centering
	\includegraphics[width=\linewidth]{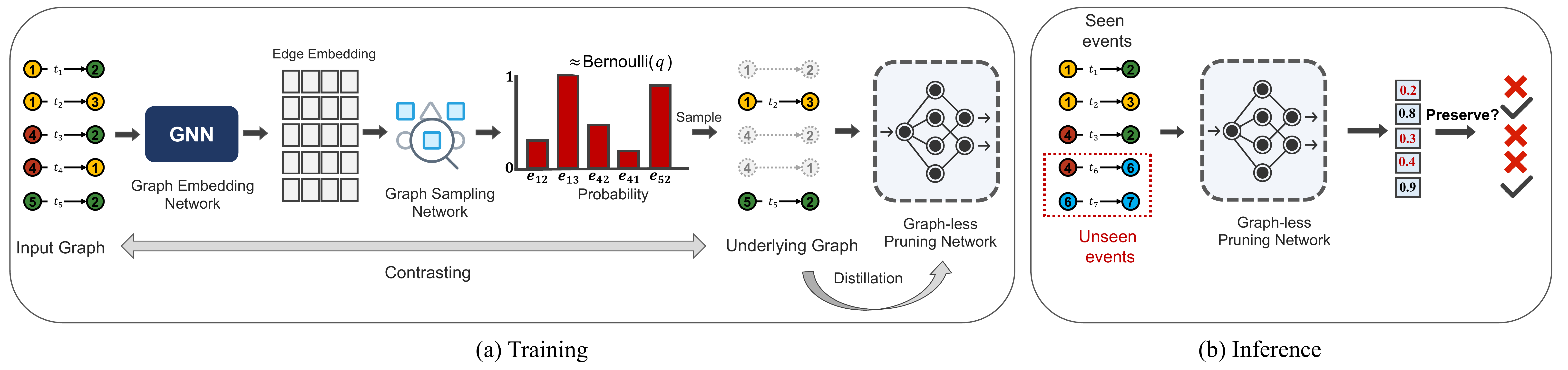}
	\caption{The overall framework of STEP. STEP consists of three main components: the graph embedding network, the graph sampling network, and the graph-less pruning network. (a) In the training stage, a graph embedding network is adopted to obtain edge embeddings of an input dynamic graph represented as a series of timed events. Then, a graph sampling network learns to produce the underlying graph as self-supervision to guide the training of the graph-less pruning network. (b) In the inference stage, the graph-less pruning network can directly deal with input temporal events and determines which edges to preserve efficiently and timely.}
	\label{fig:framework}
\end{figure*}

\subsection{Graph Pruning \& Sparsification}\label{sec:related_pruning}

Graph pruning (a.k.a. graph sparsification) aims to reduce graph storage and accelerate GNNs by constructing an underlying graph with far fewer edges. Such a graph usually preserves important structural information while retaining less noise, which would even benefit GNNs in downstream tasks. The simplest method to prune the graph is DropEdge~\cite{dropedge}, which randomly removes edges from graphs, following a specific distribution. It has been proved that a pruned graph would also help mitigate the over-smoothing issue in the training of deep GNNs.
Subsequently, NeuralSparse~\cite{neural_sparse} uses a sparsification network that learns to preserve task-relevant edges by feedback signals from downstream tasks. Furthermore, importance-based edge sampling has also been studied in a scenario where we could predefine edge importance based on explainability methods. For example, GNNExplainer~\cite{gnnexplainer} generates customized explanations by maximizing the mutual information between the distribution of possible subgraphs and the GNN's predictions. PGExplainer~\cite{pgexplainer} learns a probabilistic graph generative model to provide explanations or importance on each edge. However, most methods described in the literature mainly focus on the static graph and require access to abundant labels, which are not feasible in a practical setting.

\section{Preliminaries}
\label{sec:preliminary}
\subsection{Notations and Problem Formulation}

\subsubsection{{Notations} }
We consider a continuous-time dynamic graph (CTDG) streaming scenario, where new observations are streamed immediately and continuously. Typically, a CTDG is constructed based on plenty of temporal events $\delta(t)=(v_i,v_j, x_{ij},t)$ ordered by timestamp $t$, each event represents an interaction occurred between source node $v_i$ to target node $v_j$ at timestamp $t$; $\mathbf{x}_{ij}\in \mathbb{R}^d$ is the associated edge feature, where $d$ is the dimensionality. We denote by $\mathcal{E}=\{\delta(t_1),\delta(t_2),\ldots,\delta(t_m)\}$ consists of all temporal events in a CTDG where $m$ is the number of observed events. We denote a CTDG by $\mathcal{G}(t)=(\mathcal{V}, \mathcal{E})$, where $\mathcal{V}=\{v_i\}$ is the set of nodes involved in all temporal events. For the sake of notation simplicity, we omit the timestamp $t$ from $\mathcal{G}(t)$ hereafter.
Note that there might be more than one edge between a pair of nodes at different timestamps, so technically a CTDG is a multi-graph with numerous redundant or even noisy edges (usually pruning is necessary).

\subsubsection{{Graph pruning}}
Given an input graph $\mathcal{G}$, the main focus of graph pruning is to learn a function $\mathcal{P}: \mathcal{G} \rightarrow \mathcal{G}_s$, where $\mathcal{G}_s \subset \mathcal{G}$ is an underling graph which preserves as much informative structure of $\mathcal{G}$ as possible. Typically, one can define a pruning ratio $p (0<p<1)$ to flexibly control the number of removed edges such that $\mathcal{G}_s$ preserves $p$\% of the edges. In practice, we can manually tune the threshold of the pruning network to obtain the desired pruning ratio as well.
In the context of dynamic graphs, a pruning algorithm has to take into consideration both structural topology and time information simultaneously. Since our work focuses on the problem of \textit{unsupervised graph pruning}, no task-relevant supervision (e.g., node labels) would be provided to train the pruning network.

\section{Present Work: STEP}
\label{sec:step}
In this section, we aim to solve the \textit{unsupervised graph pruning} problem on dynamic graphs with our proposed STEP, a self-supervised temporal graph pruning framework. As shown in Figure~\ref{fig:framework}, STEP on the highest level consists of three components: a graph embedding network that produces informative node and edge representations (Section \ref{sec:embedding_network}), a differentiable graph sampling network to automatically generate the underlying structure drawn from a learned distribution (Section \ref{sec:sampling_network}), and a pruning network that can process each incoming event inductively and timely (Section \ref{sec:pruning_network}). We also detail several function losses to train the pruning network in a self-supervised fashion (Section~\ref{sec:training}).

\subsection{The Learning Objective}
\label{sec:learning_objective}

We follow the graph pruning setup in \cite{neural_sparse,pgexplainer} and adapt it to the dynamic graph modeling context. Essentially, the learning objective of graph pruning is to find the underlying graph $\mathcal{G}_s$ that preserves as many desired properties of the dynamic graph as possible, while retaining less redundant or noisy information. We approach this problem by maximizing the mutual information between the origin graph $\mathcal{G}$ and the underlying structure $\mathcal{G}_s$:
\begin{equation}
	\label{eq:MI}
	\max_{\mathcal{G}_s} \text{MI}(\mathbf{Z}, \mathbf{Z}_s) =  H(\mathbf{Z}) - H(\mathbf{Z}_s|\mathcal{G}_s),
\end{equation}
where $\mathbf{Z}$ and $\mathbf{Z}_s$ are the node representations encoded from $\mathcal{G}$ and $\mathcal{G}_s$, respectively. There exist many alternatives for mapping an input dynamic graph to low dimensional representations, such as dynamic GNNs~\cite{tgat,dyrep,jodie,tgn}.
However, it is non-trivial to directly optimize Eq.~\eqref{eq:MI} as there are $2^m$ candidates for $\mathcal{G}_s$.

To efficiently solve the objective in Eq.~\eqref{eq:MI}, we consider
each edge in the sampled discrete graph $\mathcal{G}_s$ to be drawn from a Bernoulli distribution parameterized by a graph sampling network. We relax the concrete edge distribution by a deterministic one and compute its latent variables with a parameterized embedding network. With relaxations, the sampling network is differentiable and can be optimized with gradient methods guided by self-supervisions.

\subsection{Graph Embedding Network}
\label{sec:embedding_network}
We first introduce a graph embedding network to learn the latent variables of the edge distribution.
The embedding network $f_\theta(\cdot)$ is designed as an asymmetric encoder-decoder pair, where an encoder is a function that maps from a dynamic graph to low dimensional space, followed by a decoder takes as input and produces edge embeddings of the edge $\mathbf{e}_{ij}$ for all observed events $\delta(t)$.

\subsubsection{{Node embedding}}
Since the representation of an edge depends on the embeddings of ending nodes connected together, we devise a graph encoder as a node embedding module to learn global topology information by mapping nodes to low-dimensional representations. The graph encoder takes $\mathcal{G}$ constructed at time $t$ as input to extract node representations $z_{i}(t)$ for all $v_i$ in $\mathcal{G}$. Formally, the forward propagation at $k$-th layer is described as the following:
\begin{equation}
	\label{eq:node_embedding}
	\begin{aligned}
		 & \mathbf{a}_{i}^{(k)}(t)=\operatorname{AGG}^{(k)}\left(\left\{\mathbf{h}_{j}^{(k-1)}(t): v_j \in \mathcal{N}(v_i,t)\right\}\right) \text {, } \\
		 & \mathbf{h}_{i}^{(k)}(t)=\operatorname{COMBINE}^{(k)}\left(\mathbf{h}_{i}^{(k-1)}(t), \mathbf{a}_{i}^{(k)}(t)\right),
	\end{aligned}
\end{equation}
where $\mathbf{h}_{i}^{(k)}(t)$ is the intermediate representations of node $v_i$ at the $k$-th layer. $v_j \in \mathcal{N}(v_{i} ; t)$ denotes the adjacent node with an observation $(v_i,v_j, \mathbf{x}_{ij},t_j)$ occurred prior to $t$. $\operatorname{AGG}(\cdot)$ is a differentiable aggregation function that propagates and aggregates messages from neighborhoods, such as mean or sum, followed by a $\operatorname{COMBINE}(\cdot)$ function that combines representations from neighbors and the node itself at ($k$-1)-th layer. Note that our framework allows various choices of the encoder architecture by properly designing $\operatorname{AGG}(\cdot)$ and $\operatorname{COMBINE}(\cdot)$ functions, without any constraints. We opt for simplicity and follow the practice in GraphSAGE~\cite{graphsage}. Finally, we obtain the node representations $\mathbf{z}_i(t)=\mathbf{h}_i^{(K)}(t)$ (assume that the depth of graph encoder is $K$).

\subsubsection{{Edge embedding}}
We proceed by introducing the edge decoder to learn informative edge representations that preserve the underlying structure. It is designed to capture both structural and temporal information of a dynamic graph. Given an observed event $\delta(t)=(v_i,v_j, x_{ij},t_{j}) \in \mathcal{E}$, we generate the temporal edge embedding $\mathbf{m}_{ij}$ by combining node representations associated with edge $(v_i, v_j)$, edge feature $\mathbf{x}_{ij}$, and the relative time representation as follows:
\begin{equation}
	\label{eq:edge_embedding}
	\mathbf{m}_{ij} = \mathbf{W}_e \cdot \left(\mathbf{z}_{i}\| \mathbf{z}_{j}\| \mathbf{x}_{ij}\| \Phi(\Delta t) \right) + \mathbf{b}_{e},
\end{equation}
where $\mathbf{W}_e$ and $\mathbf{b}_{e}$ are learnable parameters for the edge decoder network and $\|$ denotes the concatenation operator. For simplicity, in our notations, we omit time information $t$, which is assumed implicitly to be attached to $\mathbf{m}_{ij}$ and $z_i$.
$\Delta t=t-t_{j}$ represents the relative timespan of two timestamps for functional encodings. We use relative timespan instead of its absolute value as the former usually reveals more critical temporal information, such as the duration of an event. Correspondingly, $\Phi(\cdot)$ is a relative time encoder (RTE) that achieves a functional time projection from the time domain to a continuous space:
\begin{equation}
	\label{eq:time_encoding}
	\Phi(\Delta t) = \operatorname{Cos}\left( \mathbf{W}_{t}\cdot \Delta t + \mathbf{b}_{t}\right),
\end{equation}
where $\mathbf{W}_{t}$ and $\mathbf{b}_{t}$ are learnable parameters of RTE. This version of the time projection method follows the practice in \cite{tgat}.

\subsection{Graph Sampling Network}
\label{sec:sampling_network}
Literature~\cite{dropedge} has shown that randomly removing edges from the input graph would benefit the performance of GNNs on downstream tasks. However, such a simple method does not allow for adaptively learning the edge importance and discovering the underlying structure. In this subsection, we hereby introduce a learnable graph sampling network $S_\omega(\cdot)$ that computes the edge importance as sampling probability, and then samples sparsified subgraphs to optimize Eq.~\eqref{eq:MI}.

\subsubsection{{Measuring edge importance}}
We believe there are two main factors to consider when measuring the importance or contributions of an edge: (i) the \textit{redundancy} of edges in the original graph. (ii) the \textit{relevance} of edges to the overall graph representation.
Firstly, edge redundancy refers to the level of information redundancy of an edge compared to other edges in the graph.
Given an edge $(v_i, v_j)$, we calculate its correlation coefficients with other edges from the perspective of the attention mechanism:
\begin{equation}
	\label{eq:redundancy}
	\mathbf{s}^{rd}_{ij}  =  \sum_{(v_i,v_j)\ne (v_{i^\prime},v_{j^\prime})} \frac{ \exp( \mathbf{m}_{ij}\cdot \mathbf{m}_{i^\prime j^\prime}^\top ) }{   \sum_{(v_{i^\prime},v_{j^\prime})}  \exp( \mathbf{m}_{ij}\cdot \mathbf{m}_{i^\prime j^\prime}^\top )},
\end{equation}
where $\mathbf{s}^{rd}_{ij}$  represents the redundancy score of edge $(v_i,v_j)$, ranging from 0 to 1. A larger $\mathbf{s}^{rd}_{ij}$ indicates high redundancy and less information encoded by the learned representation $\mathbf{e}_{ij}$, and therefore the edge $(v_i,v_j)$ should be removed. 
We also introduce the edge relevance score, which refers to the amount of correlation between edge representations and downstream tasks. For example, we can maximize the cross-entropy loss of edge representations and associated labels for link classification tasks~\cite{tgat,tgn}. However, the downstream tasks are usually unaware of a pruning model and we cannot directly maximize their relevance with corresponding supervision.
Moreover, training on a single downstream task may also lead to task-specific representations with poor generalization.
These dilemmas motivate us to discover self-supervisions from the graph itself. Given that the performance on downstream tasks highly relies on the graph representations summarized from node representations~\cite{9104936}, we introduce graph pooling techniques to obtain the graph representations as downstream supervision signals:
\begin{equation}
	\label{eq:graph_representation}
	\mathbf{z}_\mathcal{G}= \operatorname{SUM}\{\mathbf{z}_i\}\ \text{or}\ \operatorname{MEAN}\{\mathbf{z}_i\}, \ v_i \in \mathcal{V}.
\end{equation}
Where $\operatorname{SUM}$ and $\operatorname{MEAN}$ are graph pooling operations that take the sum or average sum over all node embeddings.
In most cases, we can also adopt a specific node embedding $\mathbf{z}_c$ as the graph representation $\mathbf{z}_\mathcal{G}$, where $v_c$ is called the ``central node''. 
Then, we instead maximize the relevance between edge representation $\mathbf{m}_{ij}$ and graph representation $\mathbf{z}_\mathcal{G}$:
\begin{equation}
	\label{eq:relavance}
	\mathbf{s}^{rl}_{ij} =  \frac{\mathbf{W}_{rl} \cdot \mathbf{m}_{ij} \cdot \mathbf{z}_\mathcal{G}^\top}{\|\mathbf{W}_{rl} \cdot \mathbf{m}_{ij}\|\|\mathbf{z}_\mathcal{G}\|},
\end{equation}
where $\mathbf{W}_{rl}$ is a projection matrix that encodes the edge representations to match the central node representation. Here the cosine distance is used to measure the difference between edge representations and the central node representation.
Finally, the edge importance is calculated by:
\begin{equation}
	\label{eq:edge_prob}
	\mathbf{\rho}_{ij} = \mathbf{W}_s \cdot \left(\mathbf{s}^{rd}_{ij}\| \mathbf{s}^{rl}_{ij} \| \mathbf{m}_{ij}\right),
\end{equation}
where $\mathbf{W}_{s}$ is learnable parameters. $\mathbf{\rho}_{ij}$ can be also regarded as the estimated sampling probability that an edge is selected.
Then a discrete sampling function $\Gamma(\cdot)$ is implemented such that $y_{ij} = \Gamma(\rho_{i,j})$, to discover the underlying graph $\mathcal{G}_s$ that preserves useful edges while filtering potentially redundant edges. The binary variable $y_{ij} = 1$ indicates that the edge $(v_i,v_j)$ belongs to the underlying graph $\mathcal{G}_s$, and 0 otherwise.

\subsubsection{{Making samples differentiable}}
Although several sampling methods implement discrete sampling, they are not differentiable such that it is difficult to optimize the model parameters with gradient methods. To make the sampling process differentiable for training, we adopt the Gumbel-Softmax~\cite{DBLP:conf/iclr/JangGP17}  based reparameterization trick and relax the value $y_{ij}$ from a binary variable to a continuous one. The relaxation is established by assuming the graph structure follows the Bernoulli distribution, where sampling of edges from $\mathcal{G}$ are conditionally independent to each other. Following the practice in~\cite{Concrete_Distribution,neural_sparse,pgexplainer}, we utilize the binary concrete distribution to approximate the sampling process and obtain the sampled subgraph $\widetilde{\mathcal{G}}_s \approx \mathcal{G}_s $. In this way, the binarized value $y_{ij}$ is approximated according to:
\begin{equation}
	\label{eq:gumbel_softmax}
	y_{ij} = \Gamma(\rho_{ij})=  \sigma\left((\log \epsilon-\log(1-\epsilon) + \mathbf{\rho}_{ij})/\tau\right),
\end{equation}
where $\epsilon \sim \text{Uniform}(0,1)$ and  $\sigma(\cdot)$ is the sigmoid activation. $\tau$ is the temperature that controls the interpolation between the discrete distribution and continuous categorical densities. When $\tau \to 0$, ${\rho}_{ij}$ is binarized and the Gumbel-Softmax distribution resembles the discrete distribution, such that $\lim _{\tau \to 0} \widetilde{\mathcal{G}}_s = \mathcal{G}_s$.

\subsection{Graph-less Pruning Network}
\label{sec:pruning_network}
Although the graph sampling network is able to produce the pruned subgraph $\mathcal{G}_s$ drawn from a learned distribution, it cannot flexibly tackle instant message flows (e.g., a newly incoming edge) particularly when the associated nodes are unseen before.
Inspired by the success of knowledge distillation~\cite{zhang2022graphless,DBLP:journals/pami/WangY22}, we circumvent the limitation by replacing the over-parameterized graph sampling network with a compact one, graph-less pruning network, which is able to achieve high flexibility and low deployment costs.

The goal of graph-less pruning network is to filter out potentially redundant edges from a dynamic graph, similar to the graph sampling network. The key difference between them is that the graph-less pruning network requires only edge features and time information as inputs rather than the graph structure. To minimize the computational overheads and improve efficiency, we directly take a single message $\delta(t)$ as input to determine whether this message would be involved in future events. Specifically, we use a simple feed-forward neural network $g_\phi(\cdot)$ as the pruning network:
\begin{equation}
	\label{eq:pruning}
	P_{ij}= g_{\phi}(\delta(t))= \mathbf{W}_{2} \cdot \operatorname{ReLU}\left(\mathbf{W}_{1} \cdot \left(\mathbf{e}_{ij}\| \Phi(\Delta t)\right) \right),
\end{equation}
where $\mathbf{W}_{1}$ and $\mathbf{W}_{2}$ are learnable parameters of the pruning network. The pruning network directly takes an edge and related information as inputs without explicitly message passing. Compared to GNN-based alternatives, such a network is computationally efficient and is also friendly to parallel computation for large-scale processing.

\subsection{Training}
\label{sec:training}
Since the ground-truth labels are not provided during training, we cannot directly train the model with supervised loss such as cross-entropy loss. Here we instead introduce three functional losses to optimize the pruning network in a self-supervised way.

\subsubsection{{Contrastive loss}}
Inspired by the recent success of contrastive learning, we adopt InfoNCE~\cite{info_nce} as our main learning objective to maximize the representation of mutual information between the origin graph $\mathcal{G}$ and the approximated graph $\widetilde{\mathcal{G}}_s$:
\begin{equation}
	\label{eq:loss_contrast}
	\mathcal{L}_c=   -\log \frac{\exp(\mathbf{z}_{\widetilde{\mathcal{G}}} \cdot \mathbf{z}_{\mathcal{G}}  )}{ \sum_{\mathcal{G}^-} \exp(\mathbf{z}_{\widetilde{\mathcal{G}}}  \cdot \mathbf{z}_{\mathcal{G}^-})}
\end{equation}
where $\mathbf{z}_{\mathcal{G}}$ and $\mathbf{z}_{\widetilde{\mathcal{G}}}$ are graph-level representations of $\mathcal{G}$ and $\widetilde{\mathcal{G}}$, respectively.  $\mathbf{z}_{\mathcal{G}^-}$ is the graph representation of negative sample $\mathcal{G}^-$, generated by randomly removing a portion of edges from $\mathcal{G}$.

\subsubsection{{Self-distillation loss}}
Another functional loss is derived to train the graph-less pruning network $g_\phi(\cdot)$. One major barrier is how to design proper optimization that transfers knowledge from a large ``teacher'' network (i.e., graph sampling network) to a smaller ``student'' network (i.e., graph-less pruning network). In traditional distillation, it is typically implemented in two steps, that is first, to train a large teacher model, and second, to distill the knowledge from it to the student model~\cite{DBLP:conf/iccv/ZhangSGCBM19}. To further improve the training efficiency, we adopt self-distillation technique to train the graph-less pruning network by distilling knowledge within the network itself:
\begin{equation}
	\label{eq:loss_distillation}
	\mathcal{L}_s=   \frac{1}{m}\sum_{(v_i,v_j)\in \mathcal{E}}-\left(y_{ij}  \log(P_{ij}) + (1-y_{ij}) \log(1-P_{ij})\right),
\end{equation}
where $P_{ij}$ is the prediction of graph-less pruning network, as calculated by Eq.~\eqref{eq:pruning}. Here we use the approximated graph $\widetilde{\mathcal{G}}_s$ as the self-supervision (i.e., soft label) and the output of graph-less pruning network is fed into the self-distillation loss for knowledge distillation.

\subsubsection{{Regularization}}
Although we assume the graph structure follows the Bernoulli distribution and utilizes a concrete distribution to approximate the sampling process, such a constraint is not always guaranteed during training. The resulting distribution may deviate from the desired Bernoulli distribution, with numerous ambiguous values of $y_{ij}$ lying in the middle of 0 and 1. Based upon this intuition, we further incorporate an auxiliary constraint into the overall loss to enhance model training. Specifically, we propose \textit{Bernoulli Moment Matching} to fit the distribution of sampled edges by matching the distribution mean and variance with discrete probability distribution Bernoulli$(q)$:
\begin{equation}
	\label{eq:loss_bn}
	\mathcal{L}_{b} =  |\mathbb{E}[\rho] - q| + |\mathbb{E}[\rho ^2] - \mathbb{E}[\rho ]^2 - q(1-q)|,
\end{equation}
where $q$ is a prior sampling probability and $\mathbb{E}(\cdot)$ is the mathematical expectation. Recall that the mean and variance of Bernoulli$(q)$ are $q$ and $q(1-q)$, respectively. Technically, $q$ also controls the sparsity of pruned graph, similar to the pruning ratio $p$. We further justify that suitable choices of $q$ and $p$ for pruning do not degrade performance a lot in Section~\ref{sec:hypara}.

\subsubsection{{Overall loss}}
Putting them all together, the overall loss function at the training step is:
\begin{equation}
	\label{eq:loss_overall}
	\mathcal{L} =  \mathcal{L}_c + \lambda_1 \mathcal{L}_s + \lambda_2 \mathcal{L}_{b},
\end{equation}
where $\lambda_1$ and $\lambda_2$ are trade-offs for balancing the contributions of three losses.
Note that the proposed framework is not confined to dynamic scenarios and is also applicable to static graphs without time information.

\subsection{STEP: A Real-time Pruning Framework}
\label{sec:workflow}

\begin{figure}[t]
	\centering
	\includegraphics[width=\linewidth]{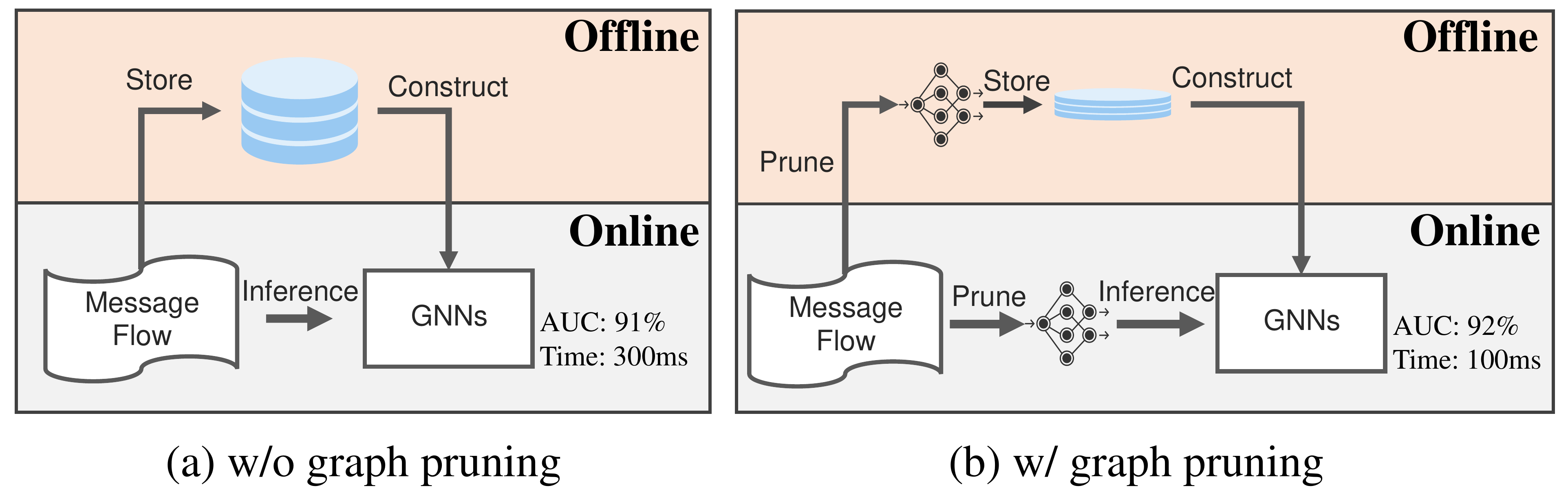}
	\caption{A typical workflow deployed on a GNN-based system to tackle streaming graph events. STEP can be incorporated into the workflow to reduce the offline graph storage and accelerates the inference of online GNNs.}
	\label{fig:workflow}
\end{figure}

As an inductive pruning framework, STEP is able to solve the problem of real-time message explosion in database storage and online inference flexibly. Specifically, we can learn the pruning network locally and do both offline and online pruning when deployed on a GNN-based system. As shown in Figure~\hyperref[fig:workflow]{3(a)}, an initial GNN-based system has to store the entire messages whenever received, resulting in massive data storage and high computational costs for message aggregation in GNNs. By contrast, STEP can significantly reduce offline storage by directly conducting pruning pre-processing on the received messages in an inductive way (Figure~\hyperref[fig:workflow]{3(b)}). In addition, the inference time and performance have also been improved as a larger number of potentially redundant messages are pruned.

\section{Experiments}
\label{sec:exp}
In this section, we report empirical evaluation results on two benchmarks (i.e., Wikipedia and Reddit) and one large-scale industrial dataset (i.e., Alipay) to validate the effectiveness and robustness of STEP framework on the dynamic node classification. More detailed descriptions about datasets, baselines, and experimental settings can be found in Appendix~\ref{appendix:setup}.
Code to reproduce our experiments is available at \url{https://anonymous.4open.science/r/PyTorch-STEP}.

\begin{table*}[t]
	\caption{Dynamic node classification task results, where the reported metric is the AUC score (\%). The best results on each dataset under different pruning ratios are boldfaced. We refer \underline{Backbone} to the standard TGN and TGAT models w/o pruning.}
	\label{tab:overall}
	\begin{tabular}{cl|ccc|ccc}
		\toprule
		\multirow{2}{*}{\textbf{Pruning Ratio $p$}} & \multirow{2}{*}{\textbf{Methods}} & \multicolumn{3}{c}{\textbf{TGN}} & \multicolumn{3}{c}{\textbf{TGAT}}                                                                                                                             \\
		\cmidrule{3-8}
		                                            &                                   & \textbf{Wikipedia}               & \textbf{Reddit}                   & \textbf{Alipay}              & \textbf{Wikipedia}           & \textbf{Reddit}              & \textbf{Alipay}              \\
		\midrule
		0                                           & Backbone                          & 85.86 {$\pm$ 0.37}               & 65.63 {$\pm$ 0.63}                & 91.54 {$\pm$ 0.71}           & 84.27 {$\pm$ 0.72}           & 68.22 {$\pm$ 0.52}           & 93.40 {$\pm$ 0.39}           \\
		\midrule
		\multirow{5}*{0.1}
		                                            & +DropEdge                         & 85.50 {$\pm$ 0.49}               & 65.35 {$\pm$ 0.27}                & 90.49 {$\pm$ 0.42}           & 84.33 {$\pm$ 0.35}           & 68.05 {$\pm$ 0.25}           & 91.64 {$\pm$ 0.28}           \\
		                                            & +NeuralSparse                     & 85.78 {$\pm$ 0.26}               & 65.62 {$\pm$ 0.36}                & 91.27 {$\pm$ 0.36}           & 84.63 {$\pm$ 0.19}           & 68.11 {$\pm$ 0.31}           & 92.48 {$\pm$ 0.27}           \\
		                                            & +PGExplainer                      & 85.51 {$\pm$ 0.31}               & 65.72 {$\pm$ 0.19}                & 91.89 {$\pm$ 0.29}           & 84.61 {$\pm$ 0.21}           & 68.41 {$\pm$ 0.17}           & 92.63 {$\pm$ 0.21}           \\
		                                            & +STEP                             & \textbf{86.06 {$\pm$ 0.25} }     & \textbf{65.90 {$\pm$ 0.21} }      & \textbf{92.39 {$\pm$ 0.23} } & \textbf{84.90 {$\pm$ 0.23} } & \textbf{69.16 {$\pm$ 0.47} } & \textbf{93.84 {$\pm$ 0.19} } \\

		\midrule
		\multirow{5}*{0.3}
		                                            & +DropEdge                         & 85.25 {$\pm$ 0.38}               & 65.45 {$\pm$ 0.22}                & 90.68 {$\pm$ 0.51}           & 84.51 {$\pm$ 0.25}           & 67.13 {$\pm$ 0.21}           & 90.17 {$\pm$ 0.30}           \\
		                                            & +NeuralSparse                     & 86.32 {$\pm$ 0.31}               & 65.47 {$\pm$ 0.27}                & 91.47 {$\pm$ 0.40}           & 85.39 {$\pm$ 0.23}           & 67.46 {$\pm$ 0.24}           & 91.07 {$\pm$ 0.26}           \\
		                                            & +PGExplainer                      & 86.96 {$\pm$ 0.37}               & 65.35 {$\pm$ 0.23}                & 91.28 {$\pm$ 0.35}           & 85.78 {$\pm$ 0.18}           & 67.55 {$\pm$ 0.19}           & 91.41 {$\pm$ 0.24}           \\
		                                            & +STEP                             & \textbf{87.27 {$\pm$ 0.23} }     & \textbf{65.62 {$\pm$ 0.32} }      & \textbf{92.18 {$\pm$ 0.27} } & \textbf{86.05 {$\pm$ 0.29} } & \textbf{68.80 {$\pm$ 0.24} } & \textbf{93.23 {$\pm$ 0.28} } \\
		\midrule
		\multirow{5}*{0.5}
		                                            & +DropEdge                         & 85.36 {$\pm$ 0.42}               & 65.83 {$\pm$ 0.30}                & 86.21 {$\pm$ 0.46}           & 84.63 {$\pm$ 0.52}           & 66.18 {$\pm$ 0.34}           & 87.11 {$\pm$ 0.32}           \\
		                                            & +NeuralSparse                     & 86.31 {$\pm$ 0.28}               & 66.06 {$\pm$ 0.35}                & 88.35 {$\pm$ 0.44}           & 85.73 {$\pm$ 0.18}           & 66.50 {$\pm$ 0.19}           & 89.86 {$\pm$ 0.36}           \\
		                                            & +PGExplainer                      & 86.40 {$\pm$ 0.36}               & 66.29 {$\pm$ 0.29}                & 89.10 {$\pm$ 0.34}           & 86.01 {$\pm$ 0.30}           & 66.69 {$\pm$ 0.25}           & 90.04 {$\pm$ 0.25}           \\
		                                            & +STEP                             & \textbf{86.73 {$\pm$ 0.28} }     & \textbf{66.63 {$\pm$ 0.34} }      & \textbf{89.88 {$\pm$ 0.31} } & \textbf{86.75 {$\pm$ 0.36} } & \textbf{68.30 {$\pm$ 0.39} } & \textbf{91.42 {$\pm$ 0.26} } \\
		\midrule
		\multirow{5}*{0.7}
		                                            & +DropEdge                         & 85.28 {$\pm$ 0.41}               & 62.65 {$\pm$ 0.42}                & 83.85 {$\pm$ 0.48}           & 84.15 {$\pm$ 0.42}           & 63.88 {$\pm$ 0.93}           & 83.20 {$\pm$ 0.38}           \\
		                                            & +NeuralSparse                     & 86.02 {$\pm$ 0.22}               & 63.37 {$\pm$ 0.39}                & 84.79 {$\pm$ 0.43}           & 85.44 {$\pm$ 0.41}           & 63.84 {$\pm$ 0.40}           & 85.34 {$\pm$ 0.41}           \\
		                                            & +PGExplainer                      & 86.24 {$\pm$ 0.29}               & 63.11 {$\pm$ 0.27}                & 85.49 {$\pm$ 0.38}           & 85.75 {$\pm$ 0.27}           & 62.28 {$\pm$ 0.32}           & 85.10 {$\pm$ 0.39}           \\
		                                            & +STEP                             & \textbf{86.35 {$\pm$ 0.31} }     & \textbf{64.42 {$\pm$ 0.33} }      & \textbf{87.51 {$\pm$ 0.29} } & \textbf{86.22 {$\pm$ 0.27} } & \textbf{64.84 {$\pm$ 0.47} } & \textbf{89.17 {$\pm$ 0.32} } \\
		\midrule
		\multirow{5}*{0.9}
		                                            & +DropEdge                         & 84.44 {$\pm$ 0.47}               & 51.20 {$\pm$ 0.43}                & 72.18 {$\pm$ 0.53}           & 83.53 {$\pm$ 0.40}           & 51.63 {$\pm$ 0.78}           & 74.94 {$\pm$ 0.47}           \\
		                                            & +NeuralSparse                     & 84.74 {$\pm$ 0.29}               & 52.89 {$\pm$ 0.42}                & 80.56 {$\pm$ 0.48}           & 84.13 {$\pm$ 0.28}           & 52.13 {$\pm$ 0.73}           & 81.03 {$\pm$ 0.44}           \\
		                                            & +PGExplainer                      & 85.19 {$\pm$ 0.30}               & 54.48 {$\pm$ 0.23}                & 80.98 {$\pm$ 0.41}           & 84.75 {$\pm$ 0.34}           & 51.82 {$\pm$ 0.20}           & 80.29 {$\pm$ 0.37}           \\
		                                            & +STEP                             & \textbf{85.21 {$\pm$ 0.30} }     & \textbf{54.83 {$\pm$ 0.32} }      & \textbf{84.32 {$\pm$ 0.34} } & \textbf{85.40 {$\pm$ 0.37} } & \textbf{53.44 {$\pm$ 0.63} } & \textbf{86.08 {$\pm$ 0.41} } \\
		\bottomrule
	\end{tabular}
\end{table*}

\begin{figure*}[t]
	\centering
	\subfigure[Wikipedia]{\includegraphics[width=0.3\hsize, height=0.2\hsize]{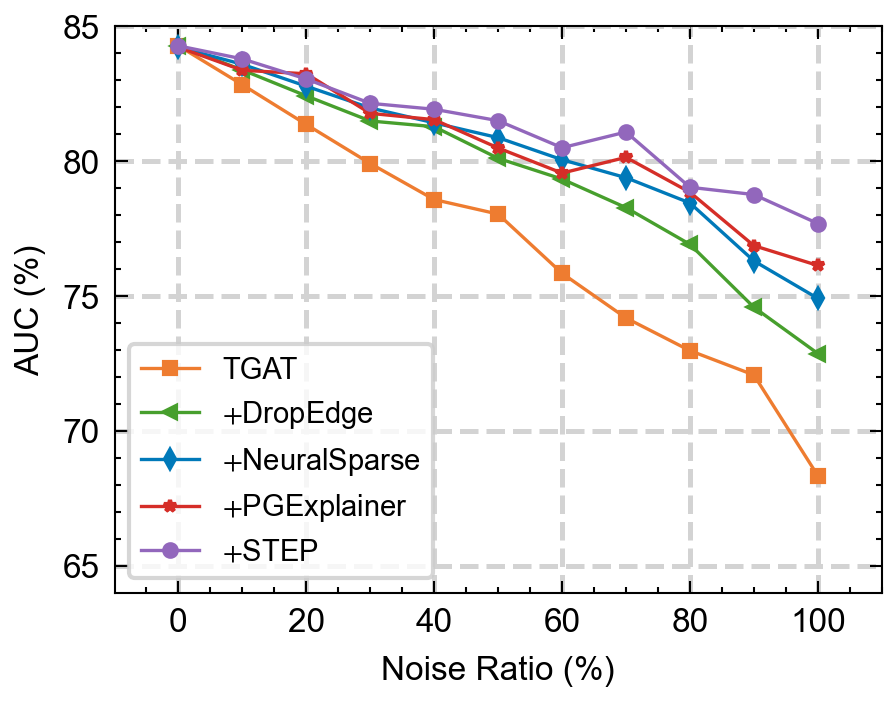}\label{fig: robustness_wiki}}\hspace{0.3cm}
	\subfigure[Reddit]{\includegraphics[width=0.3\hsize, height=0.2\hsize]{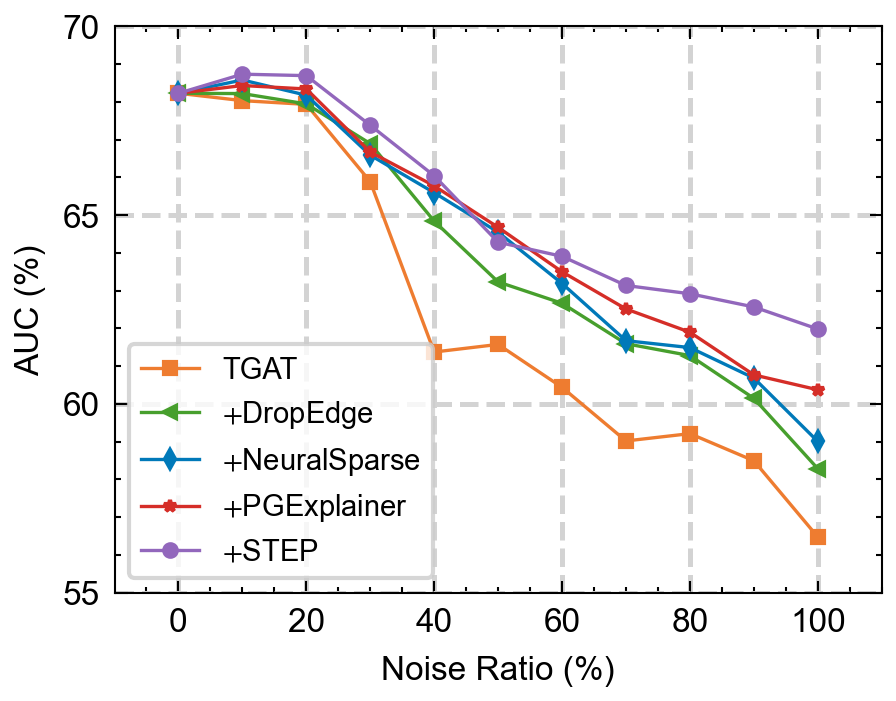}\label{fig: robustness_reddit}}\hspace{0.3cm}
	\subfigure[Alipay]{\includegraphics[width=0.3\hsize, height=0.2\hsize]{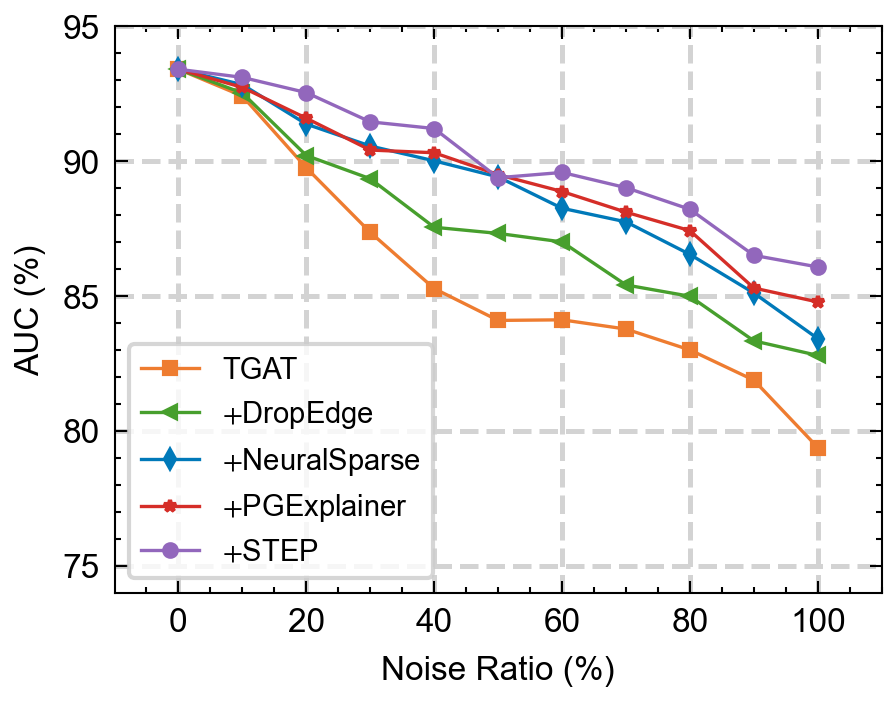}\label{fig: robustness_ali}}
	\caption{AUC score on three datasets with various levels of noise. }
	\label{fig:robustness}
\end{figure*}

\begin{figure*}[t]
	\centering
	\subfigure[Wikipedia]{\includegraphics[width=0.3\hsize, height=0.2\hsize]{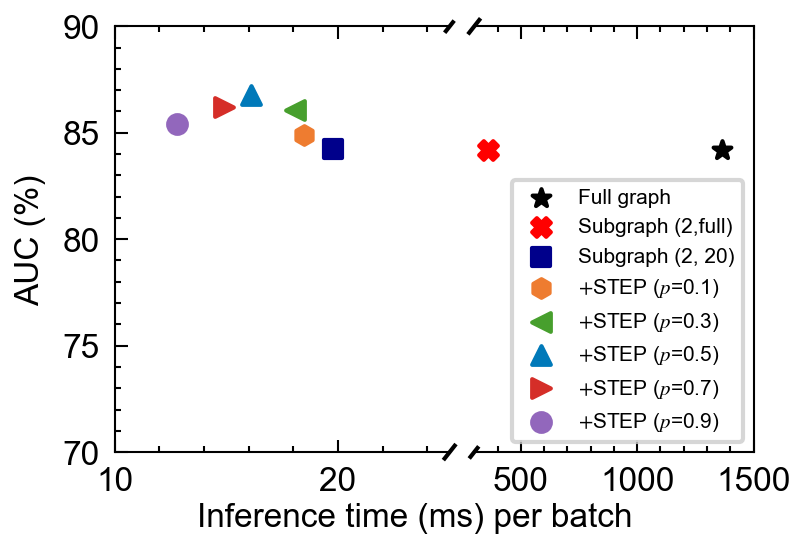}\label{fig: effe_wiki}}\hspace{0.3cm}
	\subfigure[Reddit]{\includegraphics[width=0.3\hsize, height=0.2\hsize]{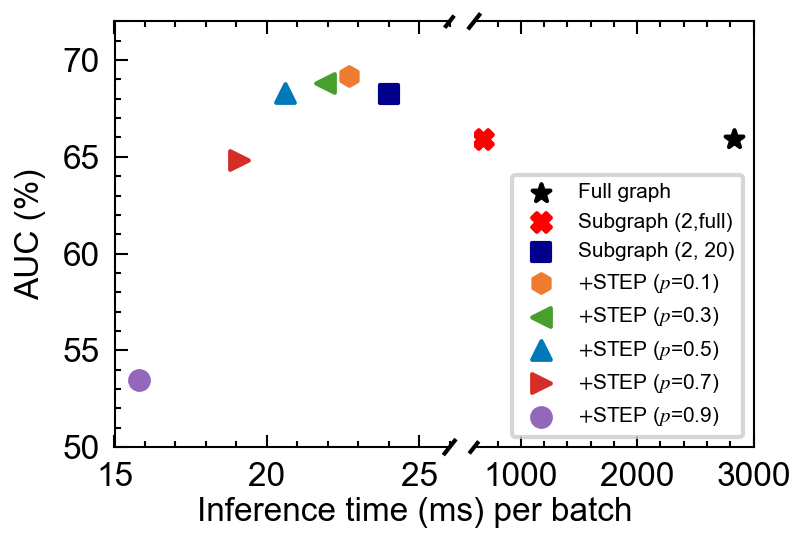}\label{fig: effe_reddit}}\hspace{0.3cm}
	\subfigure[Alipay]{\includegraphics[width=0.3\hsize, height=0.2\hsize]{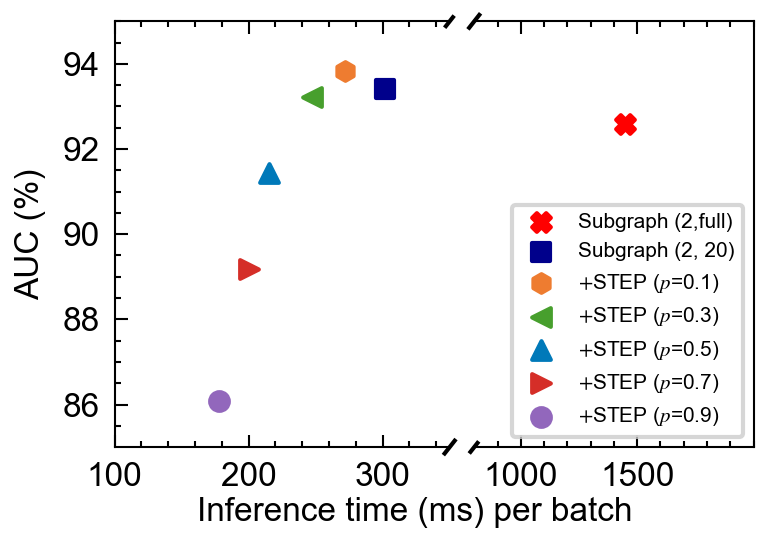}\label{fig: effe_ali}}
	\caption{ AUC score and inference time (ms) per batch of TGAT on three datasets. We omit the result of \textit{full graph} on Alipay due to ``out of memory'' error. }
	\label{fig:efficiency}
\end{figure*}

\subsection{Overall Performance}
Table~\ref{tab:overall} summarizes the dynamic node classification results on different datasets. To thoroughly evaluate the benefit of different graph pruning methods, we vary the pruning rate, i.e., the ratio of the pruned/removed edges, from 0.1 to 0.9 with a step of 0.2.

As shown in Table~\ref{tab:overall}, STEP achieves state-of-the-art performance in all cases across three datasets.
Despite the fact that no label information is provided for training, STEP still outperforms the comparison baselines.
As the graph scale enlarges, STEP can gain more advantages over baselines. Noticeably, we observe that STEP improves the performance of classification models on three datasets when the pruning ratio is set as 0.1. As the pruning ratio increases, STEP generally yields positive benefits particularly on the two relatively dense datasets Wikipedia and Reddit. The results illustrate that current graph datasets are noisy and thus removing the noisy neighbors can help nodes learn better representations.
For Alipay, a large-scale industrial dataset, we observe a performance drop in all methods when the pruning ratio is larger than 0.3. This is reasonable as it is a very sparse graph and a high pruning ratio would possibly disrupt network connectivity, thus hindering the message passing scheme of GNNs.
Nevertheless, STEP still outperforms all baselines by large margins on Alipay, and the downstream performance is not significantly sacrificed even for a particular large pruning ratio (e.g., 0.7).
This indicates that even with a very sparse graph, STEP could still learn the informative structure and preserve useful connections between nodes for downstream tasks.

Although there exists a set of noisy connections in each dataset on the dynamic node classification task, DropEdge reports slightly worse performance than the original backbones. This demonstrates that removing edges randomly cannot prune the noise edges from the graph effectively. NeuralSparse utilizes a parameterized method to actively remove potentially redundant edges according to the supervised signals. However, it constrains the sparsification network to extracted subgraphs with limited neighbors, which in return hinders its learning power and may lead to suboptimal performance in generalization~\cite{nrgnn}. PGExplainer is a strong baseline that outperforms others in most cases, which indicates the effectiveness of graph explainability methods in discovering underlying graph structure. Nevertheless, the explanations generated by PGExplainer are less accurate for tailed edges of lower importance, as evidenced by increasing $p$ would lead to poor performance and even underperforms NeuralSparse.

We can also observe that graph pruning brings more improvements for TGN than TGAT, this is because TGAT adopts self-attention mechanisms to assign different attention weights for the temporal interactions between nodes. In this way, it provides useful guidance for model training on noisy graphs. However, as the pruning ratio increase and the graph becomes more sparse, TGN combined with long-term information can lead to better performance on Wikipedia and Reddit.

\subsection{Robustness Evaluation}\label{sec:robustness}
Extensive studies have demonstrated that GNNs are not robust to noise in graph data, including inherent noise~\cite{learn_to_drop,DBLP:journals/corr/abs-2202-07114} or adversarial noise (i.e., adversarial examples~\cite{chen2020survey}).
Typically, noise is injected into the graph structure in a way that creates an additional set of noisy connections between nodes~\cite{median_gcn}.
Therefore, graph pruning would be beneficial to alleviate the negative effects of noise by removing such edges from the graphs.

In this subsection, we investigate the effectiveness of graph pruning on real-world graphs corrupted by different levels of noise. We generate noisy graphs by injecting noise edges into the original graph. The source and destination nodes of injected noisy edges are randomly selected from the node set in the original graph, and random features and timestamps (between the minimum and maximum timestamp) will be attached to each noisy edge. We use the ratio of the number of injected noise edges to the number of true edges as the independent variable for this experiment and vary the noise ratio $r$ from 0 to 1.0. The corresponding pruning ratio $p$ is set to $p = \frac{ r }{1+r}$,  in order to keep the total number of edges unchanged. We report the averaged results on three datasets with TGAT as a backbone classifier in Figure~\ref{fig:robustness}. Each experiment is run 10 times.

From the figure, we have the following observations: (i) By increasing the noise ratio, the classification performance of TGAT can significantly drop, which validates the vulnerability of GNNs against noise.
(ii) Graph pruning is an effective way to alleviate the effects of noise and also benefit the prediction.
Even by simply dropping the edge at random (i.e., DropEdge), the robustness of TGAT against different ratios of noise is improved.
(iii) Our proposed STEP consistently outperforms baselines on three datasets. With the growth of the noise ratio, the gap between our method and the baselines becomes larger.
Overall, our method is stable to yield consistent robustness improvements compared with baselines.

\subsection{Analysis of Efficiency and Scalability}
In this section, we further analyze the benefits of STEP on inference-time speed-up and storage reduction. Specifically, we conduct experiments on the full graph, a full two-hop subgraph, a sampled two-hop subgraph (20 nodes per hop), and five sampled two-hop subgraphs with different pruning ratios for each dataset. Here we omit the results of on full graph for Alipay due to computation and memory overheads.

The results are shown in Figure~\ref{fig:efficiency}. We report the overall AUC score and inference time with respect to the full graph or each batched computation graph.
The results show that: (i) Operating on a full two-hop subgraph can somehow alleviate the scalability problem of GNNs on large-scale graphs without performance sacrifices. Nevertheless, the computational overheads are still large since the localized subgraph of a node usually consists of a high number of neighboring nodes on a dynamic graph. (ii) Graph subsampling with a fixed neighborhood size can improve the model performance as well as the inference efficiency. However, the sampled subgraphs are still at scale and would additionally introduce redundancy that requires proper handling.
(iii) STEP addresses the above issues by introducing a graph-less pruning network to remove redundant computation and storage of GNNs over sampled subgraphs. As a result, it further boosts the model performance, accelerates inference, and reduces the storage given a proper pruning ratio.
(iv) STEP effectively trades off performance and efficiency, which gains higher speedups on larger graphs. As the pruning ratio increases, STEP greatly reduces the complexity and memory overheads with negligible performance loss. For an extreme case, e.g., inference on tiny edge devices that have tough resource constraints, STEP can prune 90\% of edges without significantly sacrificing the performance.

\subsection{Analysis of Training-time Pruning}
Overfitting and training instability are two major obstacles for GNNs and cannot be effectively solved by existing regularization techniques.
As shown in Section~\ref{sec:robustness}, our proposed STEP helps improve the robustness of GNNs against noise at test-time, we show that training-time pruning can also benefit the learning process of GNNs by properly removing potentially redundant edges from the noise graph.

\begin{figure}[t]
	\centering
	\centering
	\subfigure[Wikipedia]{\includegraphics[width=0.22\textwidth]{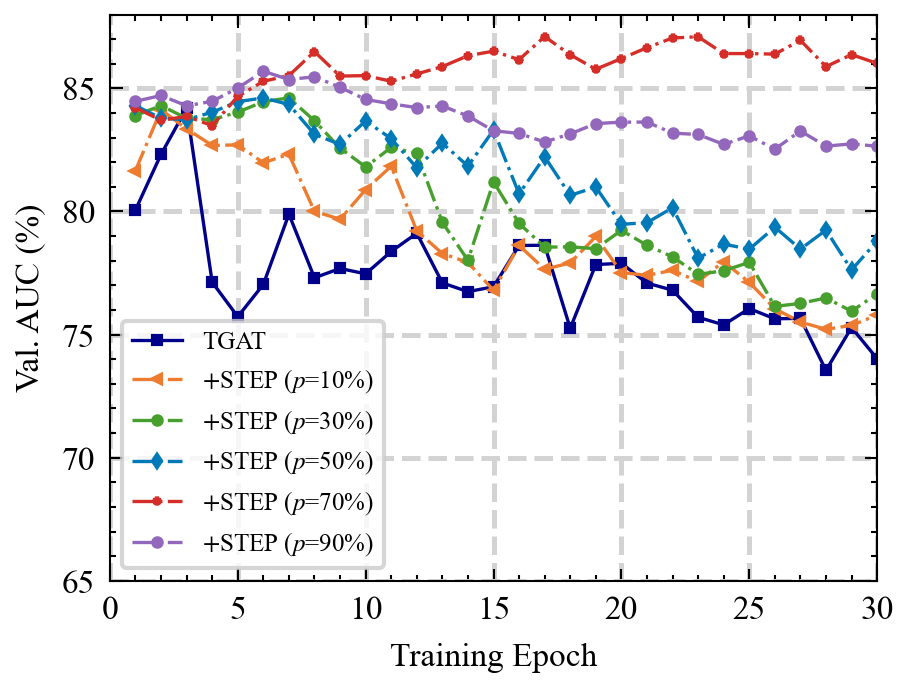}\label{fig: overfitting_wiki}}\hspace{0.3cm}
	\subfigure[Alipay]{\includegraphics[width=0.22\textwidth]{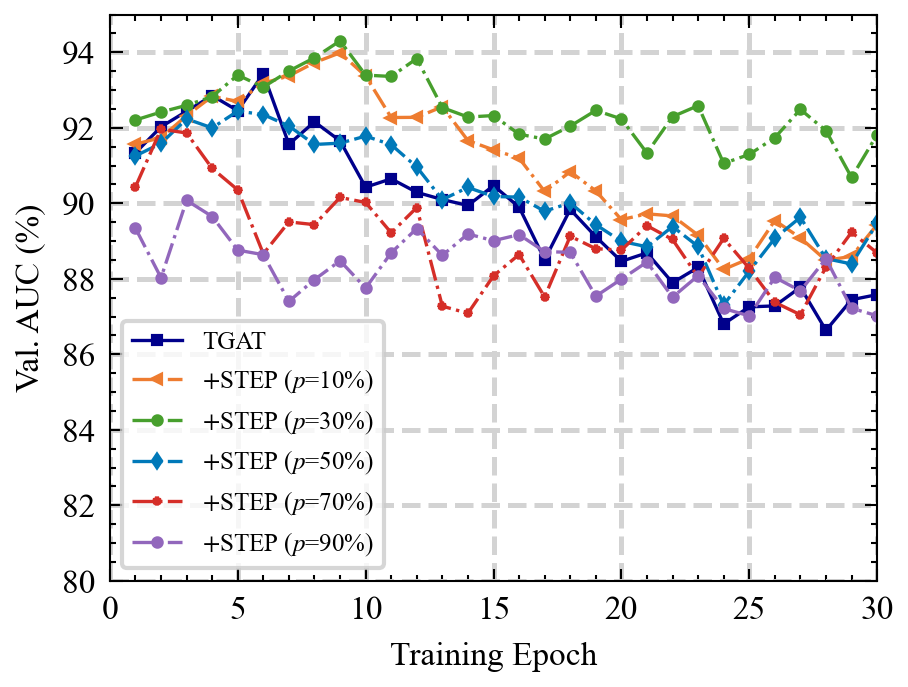}\label{fig: overfitting_ali}}
	\caption{Validation AUC convergence curve of TGAT (with STEP) on (a) Wikipedia and (b) Alipay datasets. Curves only represents the process of training phase.}
	\label{fig:overfitting}
\end{figure}

Figure~\ref{fig:overfitting} shows the validation AUC versus training epochs on Wikipedia and Alipay datasets and corresponding TGAT with various pruning ratios.
Since we have similar observations between Wikipedia and Reddit, we only report the results on Wikipedia and Alipay.
We carefully examine STEP's effects on the training phase by varying the pruning ratio from 0.1 to 0.9 with a step of 0.2.
As shown, TGAT without graph pruning is fast to be saturated on the validation set and converged to suboptimal performance on two datasets. By contrast, we utilize STEP to facilitate TGAT training and successfully boost generalization, where a consistent improvement is evident, particularly on Wikipedia. Although a large pruning ratio decreases the model performance on Alipay, it stabilizes the learning process as well.
In addition, we observe that the overfitting phenomenon is alleviated as the pruning ratio increases. Overall, STEP is able to stabilize the learning process and help GNNs achieve competitive generalization performance with sparsified graph data.

\subsection{Ablation Study}
\label{sec:abla}
In this part, we perform ablation studies to investigate the contribution of the key components in STEP, including the edge redundancy and relevance score for calculating edge sampling probability and the Bernoulli Moment Matching regularization. We obtain corresponding variants by removing functional ones from the complete STEP framework, detailed as follows: \textbf{(i)} \textbf{STEP$\backslash$(RED)} removes the part related to the redundancy score of edge sampling probability (Eq.~\eqref{eq:redundancy}); \textbf{(ii)} \textbf{STEP$\backslash$(REL)} removes the part related to relevance score of edge sampling probability (Eq.~\eqref{eq:relavance}); \textbf{(iii)} \textbf{STEP$\backslash$(RED\&REL)} removes both edge redundancy and relevance parts; \textbf{(iv)} \textbf{STEP$\backslash$(B)} replaces the Bernoulli Moment Matching regularization with a low-rank constraint $\mathcal{L}_{b} =  \sum_{(v_i,v_j}\in \mathcal{E}|  \rho_{ij} |$.

All the hyperparameters of these variants are tuned following the process described in Appendix~\ref{appendix:setup}. The pruning ratio is set as 0.5. The results of 10 runs are reported in Table~\ref{tab:abla}. Accordingly, we have the following observations: \textbf{(i)} It is clearly observed that STEP$\backslash$(RED\&REL) performs
worst on Wikipedia and Reddit, which is in line with our point that edge redundancy and relevance are two important factors in measuring the importance of edges
\textbf{(ii)} STEP$\backslash$(RED) and STEP$\backslash$(REL) perform very closely on three datasets as expected. This indicates that both the redundancy score and relevance score of edges help in increasing the quality of the pruned graph almost equally.
\textbf{(iii)} Notice that STEP$\backslash$(B) performs quite well in most experiments. However, it underperforms all other variants on Alipay, an extremely large and sparse industrial graph.
A reasonable explanation is that incorporating helpful regularization can preserve the structural information and avoid the impact of sampling bias, particularly on large and sparse graphs.
\textbf{(iv)} The complete STEP achieves the highest AUC score on all datasets, which is obviously better than other variants.
These results illustrate all the components are important in our pruning framework.

\begin{table}[t]
	\caption{Comparisons between STEP and its variants on the dynamic node classification task.}\label{tab:abla}
	\resizebox{\linewidth}{!}{
		\begin{tabular}{l|ccc}
			\toprule
			\textbf{Method}             & \textbf{Wikipedia}           & \textbf{Reddit}              & \textbf{Alipay}              \\
			\midrule
			TGAT                        & 84.27 {$\pm$ 0.72}           & 68.22 {$\pm$ 0.52}           & 93.40 {$\pm$ 0.39}           \\
			\midrule
			+STEP$\backslash$(RED)      & 85.52 {$\pm$ 0.33}           & 67.76 {$\pm$ 0.36}           & 90.87 {$\pm$ 0.28}           \\
			+STEP$\backslash$(REL)      & 85.47 {$\pm$ 0.35}           & 67.32 {$\pm$ 0.37}           & 90.69 {$\pm$ 0.27}           \\
			+STEP$\backslash$(RED\&REL) & 85.04 {$\pm$ 0.33}           & 67.05 {$\pm$ 0.39}           & 90.05 {$\pm$ 0.27}           \\
			+STEP$\backslash$(B)        & 85.67 {$\pm$ 0.41}           & 67.75 {$\pm$ 0.45}           & 89.47 {$\pm$ 0.32}           \\
			\midrule
			+STEP                       & \textbf{86.75 {$\pm$ 0.36} } & \textbf{68.30 {$\pm$ 0.39} } & \textbf{91.42 {$\pm$ 0.26} } \\
			\bottomrule
		\end{tabular}
	}
\end{table}

\begin{figure}[t]
	\centering
	\centering
	\subfigure[$p$ and $q$]{\includegraphics[ width=0.48\hsize,height=0.35\hsize]{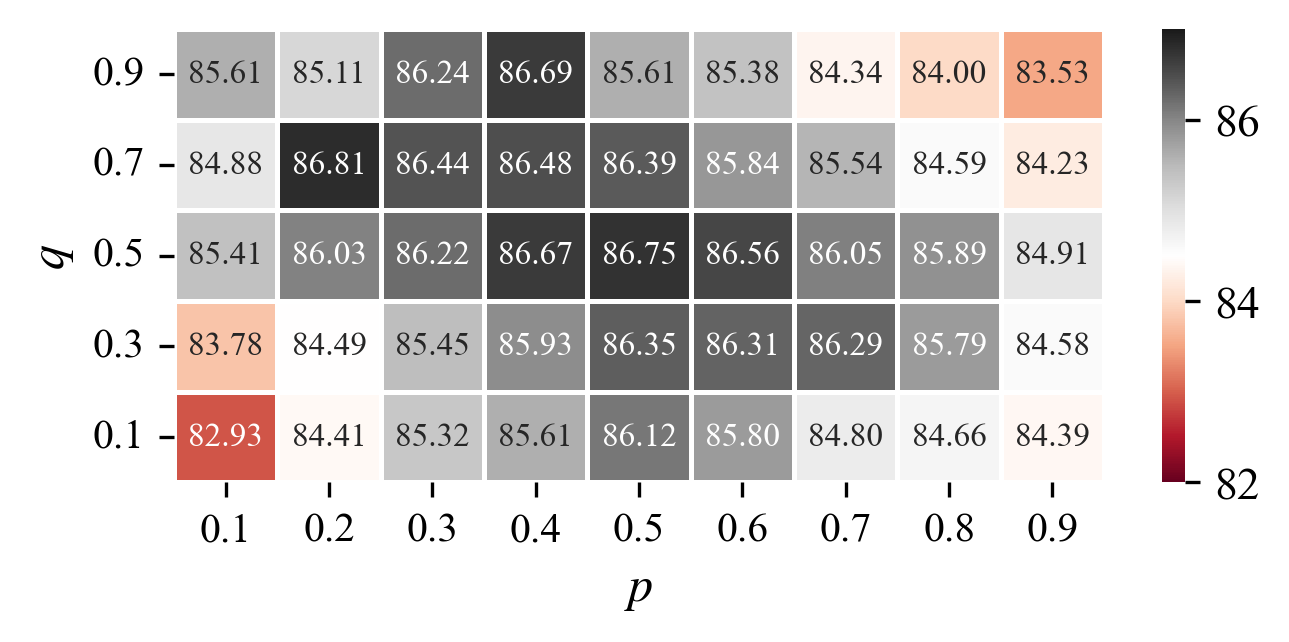}\label{fig:hypara_1}}\hspace{0.1cm}
	\subfigure[$\lambda_1$ and $\lambda_2$]{\includegraphics[width=0.48\hsize,height=0.35\hsize]{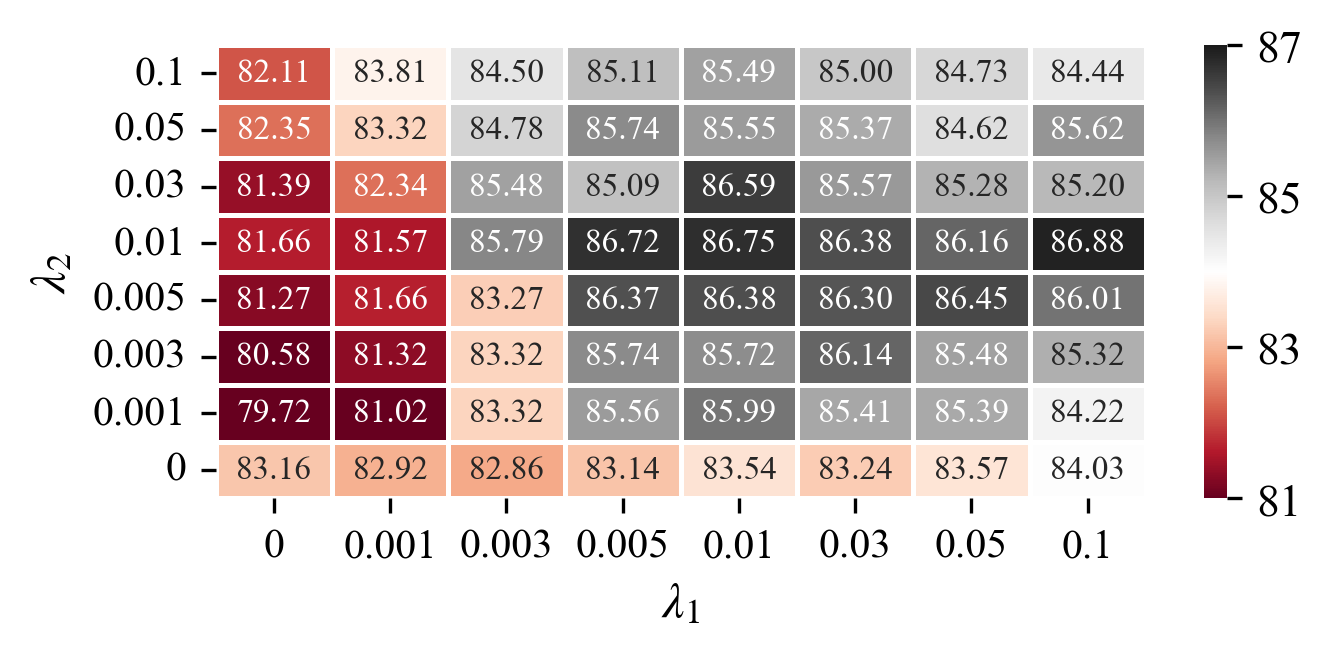}\label{fig:hypara_2}}
	\caption{Parameter analysis w.r.t. (a) $p$ and $q$, (b) $\lambda_1$ and $\lambda_2$.  We report the averaged results on Wikipedia across 10 runs.}
	\label{fig:hypara}
\end{figure}

\subsection{Hyperparameter Sensitivity Analysis}
\label{sec:hypara}
In this subsection, we first investigate how the hyperparameters $p$ and $q$ affect the performance of STEP. Specifically, $p$ controls how many edges are pruned from the initial graph, and $q$ controls the distribution of sampled edges. Both of which correspond to the sparsity of the underlying graph.
To comprehensively explore the parameter sensitivity, we vary $p$ and $q$ as \{0.1, 0.2, 0.3, 0.4, 0.5, 0.6, 0.7, 0.8, 0.9\} and \{0.1, 0.3, 0.5, 0.7, 0.9\}, respectively.  The experiments are conducted 10 times and the average results are shown in Figure~\ref{fig:hypara_1}. We make the following observations: (i) Generally, the performance of STEP tends to first increase and then slightly decrease for all $q$ as we go up to higher $p$. The performance is relatively good and stable when $p$ ranges between 0.2 and 0.8, which eases the hyperparameter selection for STEP. (ii) A similar trend can be observed on $q$, where the performance increase firstly and then decreases with the increase of $q$. We conjecture that too small $q$ would lead to a sparse graph while too large $q$ may preserve most of the noisy edges. When $q$ reaches a critical value, e.g., 0.5, the best performance is achieved in most cases.

We further study the effects of $\lambda_1$ and $\lambda_2$, two trade-off hyperparameters to balance the self-distillation loss and regularization term, respectively. As observed from Figure~\ref{fig:hypara_2}, it is clear that the pruning performance is consistently improved when $\lambda_1>0$ and $\lambda_2>0$, which means that the two terms are important for training. We can observe that STEP is sensitive to $\lambda_1$.
When $\lambda_2$ is larger we exert a stronger regularization on the learned graphs and the graphs become more sparse, which further reduces the noise information and improves the classification performance. We observe that increasing $\lambda_2$ improves the performance until the default value of $\lambda_2=0.01$ is reached. After which, the performance deteriorates. Taking together, we set $\lambda_1=\lambda_2=0.01$ as optimal values.

\section{Conclusion and Future Work}
\label{sec:conclusion}
In this work, we study the problem of unsupervised graph pruning and present a novel framework (called STEP), which aims at pruning redundant and noisy edges from a large-scale dynamic graph. By learning self-supervised signals from the input graph itself, STEP is able to discover the underlying graph that preserves the most informative structure without the feedback of task-specific supervision. Experimental results on real-world datasets show the effectiveness of the proposed framework. STEP is flexible, and can be easily incorporated into any graph-based learning pipeline.
We empirically show that the pruned graphs take much less space for offline storage and accelerate the online inference of GNNs, which are attractive to energy-efficient devices like mobile processors.
Unsupervised graph pruning is an important step toward scalable machine learning in real-world applications. In future work, we would like to extend STEP to tackle other complicated graphs, such as heterogeneous or multi-modal graphs.

\bibliographystyle{ACM-Reference-Format}
\bibliography{main}

\appendix

\section{Algorithm}
\begin{algorithm}[h]
	\caption{Training of STEP framework.}
	\label{algo:step}
	\begin{algorithmic}[0]
		\Require Graph $\mathcal{G}=(\mathcal{V}, \mathcal{E})$, embedding network $f_\theta(\cdot)$, sampling network $S_\omega (\cdot)$, pruning network $g_\phi(\cdot)$, hyperparameter $\lambda_1$, $\lambda_2$;
		\Ensure Learned pruning network $g_\phi(\cdot)$;
	\end{algorithmic}
	\begin{algorithmic}[1]
		\While{\textit{not converged}};
		\State $\mathbf{z}_i$, $\mathbf{m}_{ij} \leftarrow $ Calculate node \& edge emb.; \Comment{ Eq.~\eqref{eq:node_embedding} \& ~\eqref{eq:edge_embedding}}
		\State $\mathbf{s}^{rd}_{ij}$, $\mathbf{s}^{rl}_{ij} \leftarrow$ Calculate edge red. \& rel. score; \Comment{Eq.~\eqref{eq:redundancy} \& ~\eqref{eq:relavance}}
		\State $\mathbf{\rho}_{ij}\leftarrow$  Calculate edge sampling probability; \Comment{Eq.~\eqref{eq:edge_prob}}
		\State $y_{ij}\leftarrow$ Sample edges drawn from $\mathbf{\rho}_{ij}$; \Comment{Eq.~\eqref{eq:gumbel_softmax}}
		\State $P_{ij} \leftarrow$ Calculate edge pruning probability; \Comment{Eq.~\eqref{eq:pruning}}
		\State Calculate $\mathcal{L}_s$ according to Eq.~\eqref{eq:loss_contrast};
		\State Calculate $\mathcal{L}_p$ according to Eq.~\eqref{eq:loss_distillation};
		\State Calculate $\mathcal{L}_{b}$ according to Eq.~\eqref{eq:loss_bn};
		\State $\mathcal{L} \leftarrow \mathcal{L}_c + \lambda_1 \mathcal{L}_s + \lambda_2 \mathcal{L}_{b}$; \Comment{Eq.~\eqref{eq:loss_overall}}
		\State Update parameters by gradient descent;
		\EndWhile;\\
		\Return $g_\phi(\cdot)$;
	\end{algorithmic}
\end{algorithm}

\section{Discussions}
\subsection{Extension to Mini-batch Training}
As like many GNN alternatives, our framework is available to perform mini-batch training for large-scale graphs. Given a set of batch nodes (root nodes), we take inspiration from graph subsampling and sample their $K$-hop neighborhoods from the dynamic graph. Specifically, for a node $v_0$ at time $t_0$, we recursively sample its neighborhoods $\mathcal{N}(v_{0} ; t_0)$ up to depth $K$ to form the subgraph $\mathcal{G}^{(K)}(v_0, t_0)$. Then, the node and edge embeddings are produced by performing message aggregation on $\mathcal{G}^{(K)}(v_0, t_0)$. This matches the paradigm of current GNNs~\cite{graphsage,tgat} on large-scale graphs while alleviating the need to operate on the entire graph during training. As the $\mathcal{G}^{(K)}(v_0, t_0)$ is induced from the central node $v_0$, we thereby use $\mathbf{z}_{v_0}$ as the graph representation in Eq.~\eqref{eq:graph_representation}, rather than taking the average or sum over all node embeddings as the graph representation. Accordingly, we can simplify the negative sampling in Eq.~\eqref{eq:loss_contrast} by randomly sampling nodes from other subgraphs as negative examples.
As each node selects its edges independently from its neighborhood during subsampling, we can also utilize parallel computation on each batched subgraph to speed up the learning.

\subsection{Time and Space Complexity Analysis}
\label{sec:time_space}
Here we briefly discuss the time and space complexity of our proposed framework in terms of training and inference phases. For simplicity, we assume that the node feature, edge feature, time encoding feature, and hidden representations are $D$-dimensional. As graph sampling network and graph-less pruning network are simple feed-forward networks involved with dense matrix computation, the major computational complexity occurs in message aggregation of graph embedding network during training. Particularly, for a $K$ layer embedding network, the time complexity is related to the graph size and the dimension of features/hidden representations, about $\mathcal{O}(|\mathcal{E}|KD+|\mathcal{V}|KD^2)$. We can further incorporate mini-batch training and the time complexity for each sampled subgraph is reduced to $\mathcal{O}(|\mathcal{V}|S^KD^2)$, where $S$ is the neighborhood size shared by each hop. As for the space complexity, the major bottleneck is also the graph embedding network, which is $\mathcal{O}(|\mathcal{V}|KD+KD^2)$ and $\mathcal{O}(BS^KD+KD^2)$ for full-batch and mini-batch training, respectively. $B$ is the batch size.

At the inference stage, STEP only requires the edge feature and time information as input to compute the pruning probability $P$, the resulting time complexity is $\mathcal{O}(2LD^2)$ where $L$ is the number of layers in graph-less pruning network. Note that $P$' computation can be trivially parallelized. STEP does not introduce additional parameters besides graph-less pruning network and we do not need to store intermediate embeddings in the GPU, hence the space complexity is $\mathcal{O}(LD^2)$ for storing the network parameters. In a word, the proposed STEP is an efficient graph pruning framework available for massive graphs.

\begin{table}[t]
	\caption{ Dataset statistics. }
	\label{tab:dataset}
	\resizebox{\linewidth}{!}
	{\begin{tabular}{l|ccc}
			\toprule
			                                & \textbf{Wikipedia} & \textbf{Reddit} & \textbf{Alipay}       \\
			\midrule
			\textbf{\#Nodes}                & 9,227              & 10,984          & 7,481,538             \\
			\textbf{\#Edges}                & 157,474            & 672,447         & 21,691,814            \\
			\textbf{\#Edge features}        & 172                & 172             & 79                    \\
			\textbf{\#Labeled nodes }       & 217                & 366             & 18,330                \\
			\textbf{Density}                & 0.19               & 0.56            & 3.87 $\times 10^{-7}$ \\
			\textbf{Timespan}               & 30 days            & 30 days         & 21 days               \\
			\textbf{Positive label meaning} & posting banned     & editing banned  & fraudster             \\
			\textbf{Chronological Split}    & 70\%/15\%/15\%     & 70\%/15\%/15\%  & 14d/3d/4d             \\
			\bottomrule
		\end{tabular}
	}
\end{table}

\section{Experimental Setup}
\label{appendix:setup}
\paragraph{{Datasets.}}
We perform experiments on three real-world dynamic graph datasets, including two public datasets and one large-scale industrial dataset to validate the effectiveness and robustness of STEP framework. Datasets statistics are summarized in Table~\ref{tab:dataset}.

\begin{itemize}
	\item \textbf{Wikipedia}~\cite{tgat}. Wikipedia is a bipartite dynamic graph with $\sim$9300 nodes and $\sim$160,000 temporal edges during one month, where its nodes are users and wiki pages, and interaction edges represent a user editing a page.
	      There are only $217$ positive labels among $157,474$ transactions ($=0.14\%$), indicating whether a user is banned from posting.

	\item \textbf{Reddit}~\cite{tgat}. Reddit is also a bipartite dynamic graph containing users and interactions during one month, which contains $\sim$11,000 nodes and $\sim$700,000 temporal edges. An interaction represents posts made by users on subreddits. Dynamic labels mean whether a user is banned from posting under a subreddit. There are only $366$ positive labels among $672,447$ transactions ($=0.05\%$) in the Reddit dataset.

	\item \textbf{Alipay}. Alipay is a financial transaction network collected from Alipay platform\footnote{\url{https://www.alipay.com/}}, which consists of $\sim$7,500,000 nodes and $\sim$22,000,000 temporal transaction edges. Dynamic labels indicate whether a user is a fraudster. There are only $18,330$ positive labels among $21,691,814$ transactions ($=0.08\%$) in the Alipay dataset.
	      Note that the Alipay dataset has undergone a series of data-possessing operations, so it cannot represent real business information.
\end{itemize}
For Wikipedia and Reddit datasets, we follow the standard chronological split with 70\%-15\%-15\% according to the node interaction timestamps~\cite{tgat}. For Alipay, we adopt 14days/3days/4days split for train/val/test. Datasets statistics are summarized in Table~\ref{tab:dataset}.

\paragraph{{Baselines.}}
As the first work to study unsupervised graph pruning on dynamic graphs, we compare against three baselines adapted from the work on static graphs: (i) DropEdge~\cite{dropedge} is a heuristic method that randomly drops edges from the graphs without considering the time information. Note that DropEdge has no learnable parameters and thus no training is required. (ii) NeuralSparse~\cite{neural_sparse} utilizes downstream supervision signals to remove potentially redundant edges in an inductive manner. (iii) PGExplainer~\cite{pgexplainer} is the state-of-the-art technique that uncovers the underlying structures as the explanations and removes edges that contribute less to the downstream predictions. Both NeuralSparse and PGExplainer are supervised methods and we adapt them to the unsupervised setting by optimizing the learning objective in Eq.~\eqref{eq:MI}. For a fair comparison, we use TGN~\cite{tgn} and TGAT~\cite{tgat} as the backbone classifiers for all methods.

\paragraph{{Implementation details.}}
For all experiments, the hidden dimension of TGN and TGAT is set as 128. We report the average results with standard deviations of 10 different runs for all experiments. For the proposed STEP framework, we fix the following configuration across all experiments without further tuning:
we adopt mini-batch training for STEP and sample two-hop subgraphs with 20 nodes per hop.
The node and edge embedding dimensions are both fixed as 128. For the prior sampling probability parameter in Eq~\eqref{eq:loss_bn}, we tune $q$ over the range [0.1, 0.9], and we found that picking $q=0.5$ performs well across all datasets. We also set $\lambda_1=\lambda_2=0.01$ in Eq.~\eqref{eq:loss_overall}.
We use Adam as the optimizer with an initial learning rate of 0.001, and a batch size of 128 for the training. All parameters are carefully tuned according to the validation set performance.

\paragraph{{Software and hardware Specifications.}}
We coded our framework and the baselines in PyTorch~\cite{pytorch}. For Wikipedia and Reddit datasets, our experiments were conducted on a Linux machine with one NVIDIA Tesla V100 GPU, each with 32GB memory. While for the industrial dataset Alipay, due to high concurrent processing, we ran the experiments on distributed Linux services with 12 Intel Xeon Platinum 8163 CPUs, each with 64GB memory.

\end{document}